\documentclass[sigconf]{acmart}

\usepackage{algorithm}
\usepackage{algorithmic}
\usepackage{subfigure}
\usepackage{amsmath}
\usepackage{balance}

\usepackage{amssymb}
\usepackage{makecell}
\usepackage{multirow}

\newcommand{\norm}[1]{\left\lVert#1\right\rVert}

\newcommand*{\argmin}{\operatornamewithlimits{argmin}\limits}

\AtBeginDocument{%
  \providecommand\BibTeX{{%
    \normalfont B\kern-0.5em{\scshape i\kern-0.25em b}\kern-0.8em\TeX}}}
    
\usepackage{etoolbox}
\makeatletter
\patchcmd{\maketitle}{\@copyrightpermission}{
   \begin{minipage}{0.3\columnwidth}
     \href{https://creativecommons.org/licenses/by/4.0/}{\includegraphics[width=0.90\textwidth]{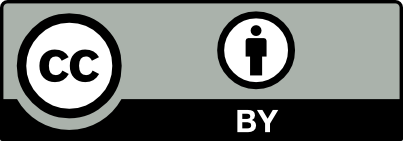}}
   \end{minipage}\hfill
   \begin{minipage}{0.7\columnwidth}
     \href{https://creativecommons.org/licenses/by/4.0/}{This work is licensed under a Creative Commons Attribution International 4.0 License.}
   \end{minipage}

   \vspace{5pt}
}{}{}

\makeatother

\copyrightyear{2022}
\acmYear{2022}
\setcopyright{rightsretained}

\acmConference[KDD '22]{Proceedings of the 28th ACM SIGKDD Conference on Knowledge Discovery and Data Mining}{August 14--18, 2022}{Washington, DC, USA}
\acmBooktitle{Proceedings of the 28th ACM SIGKDD Conference on Knowledge Discovery and Data Mining (KDD '22), August 14--18, 2022, Washington, DC, USA}
\acmISBN{978-1-4503-9385-0/22/08}
\acmDOI{10.1145/3534678.3539245}

\begin{document}

\title[Learning Differential Operators for Interpretable Time Series Modeling]{Learning Differential Operators for Interpretable \\ Time Series Modeling}

\author{Yingtao Luo}
\authornote{Work done during internship at Microsoft Research Asia.}
\affiliation{
  \institution{Carnegie Mellon University}
  \city{Pittsburgh}
  \country{USA}
  }
\email{yingtaoluo@cmu.edu}

\author{Chang Xu}
\affiliation{
  \institution{Microsoft Research Asia}
  \city{Beijing}
  \country{China}
  }
\email{chanx@microsoft.com}

\author{Yang Liu}
\affiliation{
  \institution{Microsoft Research Asia}
  \city{Beijing}
  \country{China}
  }
\email{yangliu2@microsoft.com}

\author{Weiqing Liu}
\affiliation{
  \institution{Microsoft Research Asia}
  \city{Beijing}
  \country{China}
  }
\email{weiqing.liu@microsoft.com}

\author{Shun Zheng}
\affiliation{
  \institution{Microsoft Research Asia}
  \city{Beijing}
  \country{China}
  }
\email{shun.zheng@microsoft.com}

\author{Jiang Bian}
\affiliation{
  \institution{Microsoft Research Asia}
  \city{Beijing}
  \country{China}
  }
\email{jiang.bian@microsoft.com}

\renewcommand{\shortauthors}{Yingtao Luo et al.}

\begin{abstract}
Modeling sequential patterns from data is at the core of various time series forecasting tasks. Deep learning models have greatly outperformed many traditional models, but these black-box models generally lack explainability in prediction and decision making. To reveal the underlying trend with understandable mathematical expressions, scientists and economists tend to use partial differential equations (PDEs) to explain the highly nonlinear dynamics of sequential patterns. However, it usually requires domain expert knowledge and a series of simplified assumptions, which is not always practical and can deviate from the ever-changing world. Is it possible to learn the differential relations from data dynamically to explain the time-evolving dynamics? In this work, we propose an learning framework that can automatically obtain interpretable PDE models from sequential data. Particularly, this framework is comprised of learnable differential blocks, named $P$-blocks, which is proved to be able to approximate any time-evolving complex continuous functions in theory. Moreover, to capture the dynamics shift, this framework introduces a meta-learning controller to dynamically optimize the hyper-parameters of a hybrid PDE model. Extensive experiments on times series forecasting of financial, engineering, and health data show that our model can provide valuable interpretability and achieve comparable performance to state-of-the-art models. From empirical studies, we find that learning a few differential operators may capture the major trend of sequential dynamics without massive computational complexity. 

\end{abstract}



\maketitle

\section{Introduction}

The modeling of time series data such as physiologic signals, temperature, asset price, electricity, traffic flow, and online subscribers plays an important role in various applications in our daily lives. The multivariate time series are often considered as time-evolving dynamics, where each variable depends on its historical values and other variables. The sequential trends of time series have been studied by probabilistic models for many decades, such as Gaussian process and Markov models. In recent years, deep learning models \cite{lai2018modeling, shih2019temporal, salinas2020deepar} have been intensively used for time series forecasting, outperforming previous methods in accuracy by large margins. Not only the long-short sequential patterns are considered but also the pair-wise dependencies between multivariate time series are leveraged by sophisticated learning architectures \cite{li2019enhancing, wu2020connecting}, which have successfully reduced the prediction errors. Despite the success of deep learning in various time series tasks, many real-world tasks demand stronger interpretability for the purpose of reliable and controllable decision making. Deep learning, on the other hand, raises soaring concerns about the model interpretability \cite{xing2011extracting, fortuin2019som, liu2022rmt} since the calculations are mostly conducted in the latent space, making it increasingly hard for people to understand the explicit forecasting mechanism and trust the model prediction. 

Recent advances in neural differential equations \cite{chen2018neural} revisit the use of ordinary differential equations (ODE) in modeling continuous-time dynamics with simple architectures, demonstrating a comparable performance to many deep learning models. Differential equations are widely used in quantitative subjects such as physics, biology, engineering, and finance, which reveal the inherent differential relations of variables (functions) and their rates of change (derivatives) in any local area of the dynamical system. 
Unfortunately, by using neural nets instead of explicit equations to model dynamics, neural ODE is still a black-box model and can hardly satisfy real-world's strong demand for model interpretability. Moreover, since neural ODE only considers the target variable as a univariate function of time instead of a multivariate function of other variables and time as in partial differential equation (PDE), it remains challenging to understand what variables dominate the dynamics and how do they depend on each other. 
In fact, if we can extract explicit forms of differential equations from data, such mathematical expressions can naturally provide us with knowledge about correlations between variables and how they evolve through time.

While the learning of explicit PDEs for interpretable time series modeling is tempting, two unresolved challenges need to be addressed. First, how do we represent the differential equations in explicit forms and how can we learn them from data? Although previous works \cite{dong2017image, long2018pde} have pointed out that trainable convolutional (Conv) kernels can approximate differential operators and form a PDE model for two-dimensional physical data, multivariate analyses could have much more variables (dimensions). It is computationally intractable for current methods to represent differential operations in time series. Second, many unobserved factors and unpredictable external conditions can change the dynamics. 
For example, many factors can switch the sequential pattern of stock prices, ranging from macroeconomic factors such as economic cycles to microscopic factors in investor behavior such as shifting investment preferences.
It is unlikely that a single PDE model can adapt to the ever-changing dynamics in the real world.

In this paper, we address both questions with a new learning framework that can automatically obtain interpretable differential operators from multivariate time series data. First, we propose a novel $P$-block (Polynomial-block) to capture the multivariate dynamics. It first passes the input variables through Conv layers and then fuses the output derivatives to other variables through element-wise product operation to calculate higher-dimensional differential operators. In this way, it is possible to learn the polynomial combination of target variables and any orders of derivatives to other variables and time. 
We prove that the learning of $P$-block can approximate any time-evolving continuous function in theory. 
Second, we propose a hybrid PDE model, which contains multiple PDEs that each is trained on the different time spans and sampling rates of the historical time series. A meta-learning controller is proposed to optimize the hyperparameters of this hybrid PDE model according to the current sequential pattern. This would allow the meta controller to detect and reflect the change of sequential dynamics in the time series. 

To be noted, our proposed learning of differential operators can build a hybrid PDE model that fits the time series data and learn to capture the change of sequential dynamics.
To summarize, our contributions are listed as follows:
\begin{itemize}
\item For the first time, we propose a novel $P$-block to learn interpretable partial differential equations from multivariate time series with unknown and ever-changing dynamics. 
\item We propose a meta-learning model that optimizes hyperparameters of the hybrid PDE model according to the current sequential pattern, which can effectively reflect the change of sequential dynamics for interpretability.
\item Extensive experiments on a simulated dataset and three public datasets show that the proposed method has comparable accuracy to state-of-the-art models, with clear advantages in efficiency and interpretability. 
\item We give theoretical proof that the proposed model is a universal polynomial approximator with bounded estimation error on the smoothness of time series.
\end{itemize}

\section{Related Work}
In this section, we review the existing works for sequential modeling and machine learning differential equations. 

\subsection{Time Series Forecasting}
Time series forecasting has been studied for many decades \cite{box1968some, taylor2018forecasting}. Statistical models and deep learning models are two representative schools of methods in the literature. For example, autoregressive integrated moving average (ARIMA) \cite{asteriou2011arima} is a well-known statistical model for time series forecasting. ARIMA regresses on the distribution of a multivariate variable over functions, where the data values have been replaced with the difference between their values and the previous values to handle non-stationary. The State Space Models (SSM) \cite{durbin2012time} also models time series by using a state transition function to capture the dynamics and generate observations via an observation function. These models rarely consider multivariate time series forecasting, while Gaussian Process models \cite{brahim2004gaussian} can be applied naturally to model multivariate time series data. Also, deep learning models \cite{ma2019learning, lim2021time} are rising as the new state-of-the-arts for many time series forecasting tasks. Many of them use the sequence-to-sequence neural structures \cite{sutskever2014sequence, du2018time} based on recurrent neural networks \cite{lai2018modeling, salinas2020deepar, hewamalage2021recurrent}, Convolutional neural networks \cite{bai2018empirical, hao2020temporal}, transformers \cite{li2019enhancing, wu2020deep, zerveas2021transformer}, and graph neural networks \cite{wu2020connecting}. Deep learning models are also efficient in modeling multivariate time series \cite{zhu2021mixseq}.

\subsection{Neural Differential Equations}
Neural Differential Equations represent a series of recent works that combine neural networks with differential equations to model complex dynamics. Neural Ordinary Differential Equations (Neural ODE) \cite{chen2018neural} as a family of continuous-time models is proposed to use neural networks to mimic an ODE for the dynamics and use ODE solvers to obtain the future prediction. The backpropagation can be efficiently performed by solving additional ODEs with adjoint methods. Further studies extend the capabilities of the basic neural ODE. For example, ODE-RNN \cite{rubanova2019latent} and GRU-ODE-Bayes \cite{brouwer2019gru} can model the continuous evolution of the hidden states of the RNN between two observations. Neural Jump ODE \cite{herrera2020neural} learns the conditional expectation of a stochastic process that is continuous in time. Neural controlled differential equations (Neural CDEs) \cite{NEURIPS2020_4a5876b4} are proposed to incorporate incoming information into the dynamics of ODEs. Neural stochastic differential equations \cite{tzen2019neural, liu2019neural} and neural rough differential equations \cite{pmlr-v139-morrill21b} are also analyzed. Expert knowledge is also fused into neural ODE to guide the model optimization \cite{qian2021integrating}. Although these models can capture time-dependent dynamics using univariate differentials with respect to time, the influences of other variables are not considered. 

\subsection{Dynamical System Modeling}
The modeling of dynamical systems by differential equations can date back to the early works of solving ODEs and PDEs by neural networks \cite{lee1990neural, lagaris1998artificial}. Recently, many new techniques to learn PDEs have been proposed \cite{zhu2019physics, geneva2020modeling}. For example, PDEs are reformualated as backward stochastic differential equations to apply deep neural networks to efficiently solve high-dimensional PDEs \cite{han2018solving}. Moreover, physics-informed approaches \cite{wang2020towards, karniadakis2021physics, wang2022and} are extensively studied to leverage prior knowledge of PDEs as optimization constraints to solve PDEs with sparse data. In addition, black-box neural operators \cite{li2020fourier, kovachki2021universal, lu2021learning} are proposed to learn the infinite-dimensional mapping with neural networks for nonlinear PDEs. Many efforts are also made to learn PDEs with arbitrary geometry and domain \cite{belbute2020combining, sanchez2020learning}. Except for the solving of PDEs, the scientific discoveries of PDEs by symbolic learning \cite{schmidt2009distilling, cranmer2020discovering, PhysRevResearch.4.023174} and sparse regression \cite{brunton2016discovering, rudy2017data, xu2021deep} have also been studied. To learn the nonlinear responses without fixing certain finite difference approximations, convolutional kernels are widely used as learnable differential operators \cite{long2018pde, long2019pde}. However, most works discuss PDE systems on physical data without the variation of correlations between different variables through time that changes the basic PDE structure in unforeseen ways. 

\section{Methodology}
In the following, we first give a a formal problem definition of time series forecasting and then introduce our methods.

\begin{figure*}
\centering
\includegraphics[width=0.98\textwidth]{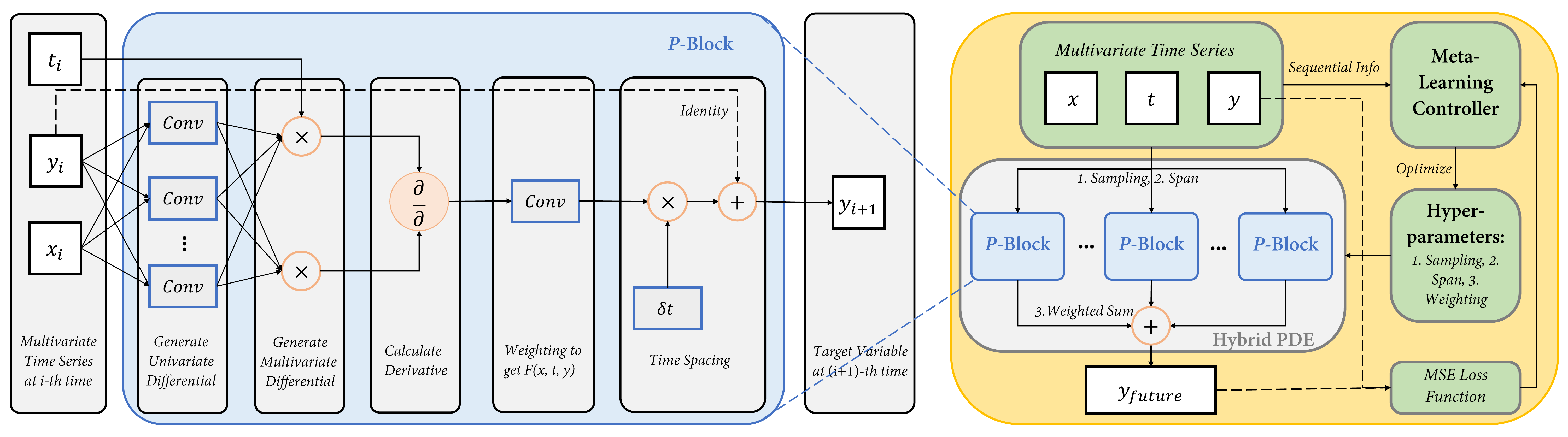}
\caption{The learning framework of $P$-block (left) and the dynamic PDE model (right). The $P$-block can take the multivariate time series as input to generate the function $F$ that parameterizes the time-evolving system dynamics, which can iteratively forecast future points via time spacing. The meta-learning controller takes various combinations of time series and hyperparameters as inputs to learn to capture the current dynamics and predict the loss. It guides us to search for the optimal hyperparameters.}
\label{fig:diagram}
\end{figure*}

\subsection{Problem Formulation}
Let us assume multivariate time series $x_{1:m} = [x_1, ..., x_{m}]$ and $y_{1:m} = [y_1, ..., y_{m}]$, in which $x_{m}$ denotes the value of the correlated multivariate time series $x$ at $m$-th time stamp, $y_{m}$ denotes the value of target time series $y$ at $m$-th time stamp, and $t_{1:m} = [t_1, ..., t_{m}]$ denotes the time stamps of the time series. We assume that each $x_{m} = [x_{m,1}, ..., x_{m,k}]$ comprises of $k$ variables and $x_{i,j}$ represents the $j$-th variable at the $i$-th time. Each $y_m$ has one variable and $y_{i}$ represents the variable at the $i$-th time. We aim to continuously predict the target variable $y_{m+1:m+\tau}$ at next $\tau$ time steps.
We are interested in the following conditional distribution
\begin{align}
    p(y_{m+1:m+\tau}|x_{1:m},y_{1:m})_{single} = \prod \limits_{i=m+1}^{m+\tau} p(y_{i}|x_{1:i-1},y_{1:i-1}),
\end{align}
In other words, whenever we have new data, we concatenate it to the historical time series and use the time series to predict the target variable for the next time. In addition, we can also consider the multi-step prediction $p(y_{m+1:m+\tau}|x_{1:m},y_{1:m})_{multi}$ directly.

Let us assume that the target variable $y=y(x,t)$ is a multivariate function of $t$ and other variables $x$. Let us further assume the response of a multivariate time series dynamical system as
\begin{align}
    \frac{\partial y}{\partial t} = F(x, t, y),
\end{align}
where $\frac{\partial y}{\partial t}$ denotes the time derivative of the target time series. Alternatively, we could also use other time differential operators such as $\frac{\partial^2 y}{\partial t^2}$, the second-order time derivative, as the left hand side of this equation.
$F$ is a function parameterizing the system dynamics with input linear, nonlinear, and differential terms to form PDE. Eq. 2 describes a dynamical system, where we assume that the time derivative is an indispensable part of the system. The equation aims at uncovering the closed-form $F$ as a partial differential equation (PDE), given time series of correlated variables and time. 

\subsection{Learnable Differential Block}
Obtaining the numerical derivatives for constructing the derivative terms $\frac{\partial y}{\partial t}$ and $\frac{\partial y}{\partial x_{:,j}}$ is crucial to the equation discovery. Traditionally, Finite Difference (FD) can calculate the derivative by dividing value differences by time differences. However, FD needs a complete ensemble of all possible differential terms according to prior knowledge. In multivariate time series, the various combinations of different variables lead to an exponentially growing number of differential terms, which makes it computationally intractable to include all possible terms. To handle this issue, we introduce a learnable differential block, namely, the $P$-block. The architecture of the $P$-block is shown in Figure \ref{fig:diagram}. The input state variable to $P$-block would firstly be mapped to the hidden space through multiple parallel Conv layers, whose output is then fused via the element-wise product operation. A Conv layer with $1 \times 1$ kernel \cite{lin2013network} is subsequently used to linearly combine multiple channels into the output to approximated $F$. The use of Conv layers is to approximate the differential operators as previous works \cite{dong2017image, long2018pde} point out and replace the FD with a learnable differential method. With learnable differential blocks, we can obtain the differential operators via training without assuming a term library to the system a priori. Here, a $P$-block seeks to approximate the function $F$ via polynomial combinations of target variable $y$ and its differential terms, given by
\begin{align}
    F(x,t,y) = \sum_{c=1}^{N_c} f_c \cdot [\prod_{l=1}^{N_l} \prod_{j=1}^{k} (K_{(c,l)} \circledast y) / (K_{(c,l)} \circledast x_j)T],
\end{align}
where $N_c$ and $N_l$ denote the number of channels and parallel Conv layers respectively. $\circledast$ denotes the Conv operation with $N \times 1$ kernel. $(c, l)$ indicates the Conv filter of $K$ of $l$-th layer and $c$-th channel. $f_c$ denotes the weights in $1 \times 1$ Conv layers with bias terms omitted. $T=Dropout(t)$ can introduce time into the system when needed. The proposed block only calculates the division of $y$ differentiation by $x$ differentiation of subsequent time steps, thus it can work on different distributions of time series. 

The diagram of the proposed learnable differential block is shown in Figure \ref{fig:diagram}. First, the target variable $y$ and correlated variables $x$ are processed by the differential Conv operators $K_{(c,l)}$ as shown in Eq. 3. This allows the system to create differential terms such as $\partial y$ and $\partial x_{:,1}$. Then, these terms are processed by two element-wise product operations together with $t$ to create multivariate polynomials of differential terms for both $y$ and $x$, such as ($\partial y^2$), ($x_{:,2} \cdot \partial x_{:,1}$), ($t \cdot \partial^2 y$). In Eq. 3, the inner product operation $\prod_j$ can multiply differentials of different variables to create multivariate 
differentials, and the outer product operation $\prod_l$ can multiply the same differentials multiple times to create higher-order terms. Afterwards, by the division operation and weighting, we can obtain the additive combination of multiple nonlinear polynomial derivatives like $t \cdot \frac{\partial^2 y}{\partial x_{:,1} \partial x_{:,2}} + \frac{\partial^2 y}{\partial x_{:,2}}$. 

Due to the learnable property of the proposed $P$-block, we can uncover the most fitted differential terms into the PDE without constructing a large ensemble of differential terms a priori. Limited to the pair-correlation of the variables in the multivariate time series, one Conv layer can only approximate a derivative of one variable. However, the proposed element-wise product operations make it possible to multiply univariate derivative terms to form multivariate derivative terms. By $P$-block, we can adapt the modeling of the dynamical system to high-dimensional multivariate time series. 

\subsection{Sparse Regression}
Based on the proposed $P$-block to calculate $F$ in Eqs. (2-3), we use the sparse regression algorithm to optimize the model and make sure the obtained $\hat{F}$ has a concise form. The motivation to have a sparse representation of $\hat{F}$ is to maintain an interpretable mathematical expression of the PDE. Without this constraint, the discovered PDE model will take up an enormously large equation space with many derivatives in the function. Moreover, when the orders of derivatives are too high, the differential accuracy also drops drastically, which adds extra errors to the model. Therefore, the sparse regression roots on the observation that only a few terms can reflect the major trend, and many other terms that are not influential enough are often considered trivial. Let us denote the number of terms in $F$ as $n$, we have $\frac{\partial y}{\partial t} \in R^{m \times 1}$. We can concatenate all the candidate terms in $F$ and form a candidate matrix $X \in R^{m \times n}$. We can denote the sparse coefficient vector as $W \in R^{n \times 1}$ and reformulate Eq. 2 to a sparse representation of the dynamical system as
\begin{align}
    \frac{\partial y}{\partial t} = XW.
\end{align}
To compute a proper $W$ so that it is both accurate in regression and sparse enough to infer interpretable mathematical insights, here we use the accelerated proximal gradient (FISTA) approximation of LASSO regression. The overall optimization of the coefficient vector accords to the following criteria:
\begin{align}
    W^* = \argmin_{W} \norm{\frac{\partial y}{\partial t} - XW}_2^2 + \lambda \norm{W}_1,
\end{align}
where $\norm{W}_1$ measures the sparsity of coefficient vector, i.e. the sparsity of differential terms in the PDE. $\lambda$ is another coefficient to balance the regression errors and sparsity, which can be adjusted flexibly due to the preference of sparsity or accuracy.  

\subsection{Recurrent Differential Model}
The proposed $P$-block is used to construct the nonlinear function $F$ in Eq. 2. How to utilize the proposed block for time series forecasting requires a comprehensive neural architecture. In neural ODE, such forecasting function is undertaken by a black-box ODE solver. However, directly solving PDEs can be much harder than ODEs, which calls for a different approach. Herein, we propose a recurrent differential model to use $P$-block for inference, which mimics the forward Euler time stepping 
\begin{align}
    y_{i+1} = y_i + \hat{F}(y;\omega) \delta t,
\end{align}
where $\delta t = t_{i+1} - t_{t}$ denotes the time difference of the two adjacent target variables in the time series. $\omega$ denotes the trained parameters in the $P$-block to approximate $F$. As shown in Figure \ref{fig:diagram}, the time stepping together with the identity mapping of previous $y_i$ can obtain the estimated $y_{i+1}$ to realize Eq. 6. For the multi-step time series, we can stack one $P$-block for multiple times for $y_i \rightarrow y_{i+1}$, $y_{i+1} \rightarrow y_{i+2}$, etc., as shown in Figure \ref{fig:diagram}. In this way, we can use $F$ as a nonlinear differential operator on the variables to mimic each time's variation throughout the entire time series and make future predictions. It should be noted that the accuracy of approximating $F$ is at the core of reducing forecasting errors, which will accumulate when stacking predicted future time series to historical time series and further making multi-step predictions. 

\subsection{Hybrid and Dynamic PDE Model with Meta-learning Mechanism}
Many observational studies have informed us that the sequential dynamics of most time series are not fixed, but have an ever-changing pattern. When some unobserved nontrivial variables or some external conditions change, the dynamics will change unpredictably. This phenomenon is sometimes referred to as concept drift, or out-of-distribution shift, meaning that the test set does not necessarily share the same data distribution with training data. To tackle this issue, we first assume that the dynamics within a shorter period are generally smoother than the dynamics of the longer period. In fact, non-stationary processes are usually treated as stationary in local areas when only a slight time difference $dt$ is considered. On the other hand, the change of external conditions or unobserved variables could have inherent patterns. For example, the periodic patterns such as the last trading day of each quarter for the stock markets have an impact on the change of dynamics. Based on the two facts, it seems wiser to design a mechanism by which we can detect the dynamics change and use different strategies for time series forecasting. Unlike neural networks with long-short memory mechanisms, a PDE as a single equation is not designed to balance the global-local dynamics. Also, it cannot capture dynamic change and make robust predictions under this scenario. 

In this paper, we propose a hybrid PDE model to improve the performance of PDE models. First, we create $h$ reorganized time series by hyperparameters time spans $s = [s_1, ..., s_h]$ and sampling rates $r = [r_1, ..., r_h]$. By time span $s_i$, we can extract a proportion of data $[x_{m-s_i : m}]$ and $[y_{m-s_i : m}]$ from the original time series; by sampling rate $r_{i}$, we can extract $[x_{1}, x_{1+r_{i}}, x_{1+2r_{i}}, ...]$ and $[y_{1}, y_{1+r_{i}}, y_{1+2r_{i}}, ...]$. Combining the two hyperparameters, we can obtain $x^{(i)} = [x_{m-s_i}, x_{m-s_i+r}, ...]$, $y^{(i)} = [y_{m-s_i}, y_{m-s_i+r_{i}}, ...]$ and $t^{(i)} = [t_{m-s_i}, t_{m-s_i+r_{i}}, ...]$. We propose a hybrid PDE model that consists of multiple PDEs as an weighted ensemble model to predict the time derivative, which reformulate Eq. 2 to 
\begin{align}
    \frac{\partial y}{\partial t} = H(x, t, y, s, r, \epsilon) = \sum_{i=1}^h \epsilon_i F_i(x^{(i)}, t^{(i)}, y^{(i)}),
\end{align}
where $\epsilon_i$ is the weight coefficient, a normalized hyperparameter scaling the contribution of each PDE in the hybrid model. $\sum_{i=1}^h \epsilon_i = 1$. $H$ represents the prediction of time derivative of the hybrid model, in contrast to $F$. In essence, each $F_i$ is independently trained with different time series such that we can find the optimal long-short balance for the data distribution. 

Furthermore, we propose a meta-learning model to capture the dynamics change and update the hyperparameters in the hybrid model dynamically. We hope to efficiently optimize the hyperparameters in real time for every prediction, instead of tuning the hyperparameters once for the entire dataset. To this end, we propose to extract latent states from the time series that reflect the change of dynamics to guide the optimization of hyperparameters in the hybrid model. We can obtain a meta-controller $G$ by 
\begin{align}
    \argmin_{\theta} G(x, t, y, s, r, \epsilon; \theta) - H(x, t, y, s, r, \epsilon; \omega) + \frac{\partial y}{\partial t},
\end{align}
where $\omega$, i.e. the model parameters of the hybrid PDE, is fixed, and we optimize $\theta$, i.e. the model parameters of the meta-learning controller. 
$G$ seeks to predict the estimation error of the hybrid PDE model $H$ by time series information and hyperparameters. It is trained by minimizing the regression error of $\frac{\partial y}{\partial t} - H$ under various combinations of time series and hyperparameters. The training process of $G$ is shown in Algorithm 1 in Appendix. When the training completes, we can use $G$ for inference to quickly search for the optimal hyperparameters in real time whenever we have new data coming into the time series. In our paper, $G$ is set to be the combination of an LSTM model processing the latent states of the time series and an MLP processing the hyperparameters. 

\section{Theoretical Analysis}
First, we prove that the proposed $P$-block is a universal polynomial approximator, meaning that the proposed block can approximate any polynomials of nonlinear differential operators. To support this claim, we need to recognize that $P$-block roots on the numerical differentiation (i.e., forward Euler time stepping), which can guarantee the convergence of the solution under two conditions: (i) the right-hand side of Eq. 1, i.e. $F$, can be estimated accurately; and (ii) the time spacing $\delta t$ is sufficiently small. The first condition can be satisfied by the optimization of sparse regression stated by Eq. 5. The second condition can
be satisfied by selecting an appropriately small $\delta t$ to reflect the time derivative. 

The design of $P$-block is essentially the flexible combination of additive representations and multiplicative representations of differential operators. With these two representations, we can express any polynomials of differential operators that we construct on the observable variables, such as $\frac{\partial^2 y}{\partial x^2_{:,2}}$, $t \cdot \frac{\partial^2 y}{\partial x_{:,1} \partial x_{:,2}}$.
This can be argued by proving that each monomial can be approximated by multiplying the output of a given number of Conv layers. To this end, we provide the following theoretical proof reproduced from \cite{long2019pde}.

For $N \times 1$ Conv filter $q$, define the vector of $q$ as
\begin{align}
    V(q) = (v_i)_{N \times 1},
\end{align}
where
\begin{align}
    v_i = \frac{1}{i!} \sum_{g=-\frac{N-1}{2}}^{\frac{N-1}{2}} g^i q[g], i=0, 1, ..., N-1
\end{align}
By Taylor’s expansion, we can formulate the convolution on any smooth function $f$, i.e. the differential operator on $f$, as
\begin{align}
    L(y) &= \sum_{g=-\frac{N-1}{2}}^{\frac{N-1}{2}} q[g] f(x+g \delta x) \nonumber\\
    &= \sum_{g=-\frac{N-1}{2}}^{\frac{N-1}{2}} q[g] \sum_{i=0}^{N-1} \prod_{j=0}^{k} \left.\frac{\partial^i f}{\partial^i x_j}\right|_{x} \frac{g^i}{i!} \delta x^i + o(\sum_{j=0}^{N-1} |\delta x_j|^{N-1}) \nonumber\\
    &= \sum_{i=0}^{N-1} v_{i} \delta x_i \cdot \prod_{j=0}^{k} \left.\frac{\partial^i f}{\partial^i x}\right|_{x} + o(\sum_{j=0}^{N-1} |\delta x_j|^{N-1}).
\end{align}

From Eq. 11, we can conclude that filter $q$ can approximate any differential operator with the prescribed order of accuracy by imposing a constraint on $V$. Without loss of generality, if we consider a sufficiently large $N$, the proposed $P$-block is a universal approximator to polynomial functions. With this relaxation, we show that the proposed $P$-block can approximate any time-evolving continuous function in theory by the generalized Stone–Weierstrass theorem \cite{stone1948generalized}. The estimation error of $P$-block is bounded by the minimal terms related to the variable value difference $\delta x$. Therefore, for time series that are generally smooth, $P$-block can accurately approximate the desired differential operators. A special case of $P$-block is that if the weights in the convolutional kernels are all fixed to one, the model only calculates the polynomials of linear and nonlinear terms without differential operators. 

\section{Experiments}
In this section, we conduct extensive experiments on simulated data and three real-world data for time series forecasting tasks.


\subsection{Datasets}
We evaluate our proposed differential equation learning method on a simulated dataset and three real-world datasets. 
\begin{itemize}
\item \textbf{Synthetic Dataset} We use a complex polynomial function to simulate complex time series. Consider $y$ as target variable, $x_1$ and $x_2$ as correlated variables, and $t$ as time, the equation we use is 
\begin{align}
    y(x_1, x_2, t) = \frac{16}{\pi^2} \sum_{k,l=1}^{40} \frac{[(-1)^k-1][(-1)^l-1]T}{k^3l^3};  \\ 
    \text{where} \quad T = sin(k\pi x_1)sin(l\pi x_2)cos(\pi t \sqrt{k^2+l^2}). \nonumber
\end{align}
We generate 1000 points within [0,10] for $t$, 1000 points for $x_1=|cost|^2$, $x_2=|sint|^2$ and $y(x_1, x_2, t)$. This function borrows from a special solution of a wave equation \cite{langer1937connection} in physics as $\frac{\partial^2 y}{\partial^2 t} = \frac{\partial^2 y}{\partial^2 x_1} + \frac{\partial^2 y}{\partial^2 x_2}$. We apply it to time series scenario here to mimic a complex dynamics.

\item \textbf{Orderbook Dataset} The Orderbook dataset contains high-frequency trading orderbooks collected from real stock markets. The time interval for futures is 5 seconds and the time interval for spot prices is 500 milliseconds. We re-sample the spot prices with a 5s interval that matches the times of futures prices. Every day, there are three and a half hours for trading. We stack a year's data sequence and reorganize the time info from the first second using the 5-second interval. We use the tick price of futures as the target variable and treat the price of the stock market as a correlated variable. We perform a moving average on both future and stock prices (5s-level) of every 500 points.

\item \textbf{MISO Dataset} We use actual energy price and actual load in the year 2021 from the Midwest Independent Transmission System Operator (MISO) database\footnote{\url{http://www.energyonline.com/Data/GenericData.aspx?DataId=8&MISO}}, open-access transmission service and monitoring the high-voltage transmission system. There are 104179 data points in the actual load time series and eight regional hubs containing 42377 data points in total in actual energy price. The average sampling rate is every 5 minutes for load and every 1 hour for price. We regard load as target variable and treat prices as correlated time series.

\item \textbf{PhysioNet Dataset} We use the PhysioNet/Computing in Cardiology Challenge 2012 dataset \footnote{\url{https://www.physionet.org/content/challenge-2012/1.0.0/}}, an open-access medical dataset that contains 8000 time series from ICU stays, each containing measurements from the first 48 hours of a different patient's admission to ICU. All patients were adults who were admitted for a wide variety of reasons to cardiac, medical, surgical, and trauma ICUs. Measurements were made at irregular times and there are 37 variables in the time series. We will predict respiration rate while regarding others as possible correlated variables.
\end{itemize}

Comparatively, the Synthetic dataset has the highest sampling rate as the smoothest time series, as shown in Figure \ref{syn}. The Orderbook dataset is the moving average of high-frequency futures and spot prices, which is also fairly smooth in local areas. MISO dataset has a very low sampling rate of data and PhysioNet has an irregular sampling of data, which makes the two datasets harder to model. Another big challenge for time series forecasting on real-world datasets is that we are not fully exposed to all the correlated variables but only have limited observable variables. For example, we do not know of the global events that may affect the stock market and the nutrient intake from food by patients as external conditions.

\subsection{Baseline Models}
We compare our method to the following baselines for evaluation.
\begin{itemize}
\item \textbf{LSTM:} \cite{hochreiter1997long} A recurrent neural network with long-short term gating mechanism.
\item \textbf{ARIMA:} \cite{box2015time} The autoregressive integrated moving average model for time series forecasting.
\item \textbf{Prophet:} \cite{taylor2018forecasting} A classical forecasting procedure based on an additive model.
\item \textbf{Neural ODE:} \cite{chen2018neural} A deep neural network based on black-box ordinary differential equation solver.
\item \textbf{ODE-RNN:} \cite{rubanova2019latent} A recurrent neural network with black-box ordinary differential equation solver.
\item \textbf{ConvTrans:} \cite{li2019enhancing} A state-of-the-art transformer model with enhanced locality capability by convolutional layers.
\item \textbf{DeepAR:} \cite{salinas2020deepar} A state-of-the-art probabilistic autoregressive recurrent networks.
\end{itemize}

\begin{table*}[t]
\centering
  \caption{Performances on different models for multi-step time series forecasting in test RMSE. Best performances are indicated by bold fonts and the strongest baselines are underlined. RMSE times the modifier is authentic RMSE.}
  \label{tab:main1}
\begin{tabular}{ccccccccccccc}
\toprule
Dataset & Modifier & LSTM & ARIMA & Prophet & Neural ODE & ODE-RNN & ConvTrans & DeepAR & Ours\\
\midrule
Synthetic   &   $\times 10^{-8}$   & 5.0148      & 5.1317        & 5.0192      & 4.9622
            & 4.9454       & 4.8586        & \underline{4.8149}       & \textbf{4.7933}\\
Orderbook   &   $\times 10^{-6}$   & 2.8960      & 6.0201        & 4.0005       & 2.4719
            & 2.5523       & \underline{0.9690}        & 1.6672       & \textbf{0.8795}\\
MISO        &   $\times 10^{-2}$       & 5.4411       & 5.2085        & 5.1413       & 5.2085
                & 4.7990       & \underline{4.0583}        & 4.2171       & 4.1489\\
PhysioNet   &   $\times 10^{-3}$   & 3.6438      & 3.5597        & 3.2567       & 2.4400
              & 2.3551      & \underline{2.3291}        & 2.4673       & \textbf{2.3266}\\
\bottomrule
\end{tabular}
\end{table*}

\begin{table*}[t]
\centering
  \caption{Performances on different models for  single-step time series forecasting in test RMSE. Best performances are indicated by bold fonts and the strongest baselines are underlined. RMSE times the modifier is authentic RMSE.}
  \label{tab:main2}
\begin{tabular}{ccccccccccccc}
\toprule
Dataset & Modifier & LSTM & ARIMA & Prophet & Neural ODE & ODE-RNN & ConvTrans & DeepAR & Ours\\
\midrule
Synthetic   &   $\times 10^{-8}$   & 4.9727      & 5.0254        & 5.0124      & 4.9030
            & 4.8862       & 4.6472        & \underline{4.6036}       & \textbf{4.4829}\\
Orderbook   &   $\times 10^{-9}$   & 3.2354      & 3.4782        & 4.0104      & 2.6025
            & 3.0146       & \underline{2.4399}        & 2.7490       & \textbf{1.5495}\\
MISO        &   $\times 10^{-2}$       & 5.2237       & 5.0837        & 4.9920       & 4.7274
                & 4.5962       & \underline{3.8489}        & 4.0311       & \textbf{3.6263}\\
PhysioNet   &   $\times 10^{-3}$   & 3.2853      & 3.2161        & 3.1566       & 2.3555
              & 2.3282      & \underline{2.1460}        & 2.1805       & \textbf{2.0260}\\
\bottomrule
\end{tabular}
\end{table*}

\begin{table}[t]
\centering
  \caption{Performances on our model for the multi-step time series forecasting in test RMSE w/o meta-learning controller.}
  \label{tab:main3}
\begin{tabular}{ccccc}
\toprule
Dataset & Modifier & Meta & Hybrid & Single\\
\midrule
Synthetic   &   $\times 10^{-8}$     & \textbf{4.7933}       & 4.9284        & 4.9581\\
Orderbook   &   $\times 10^{-6}$      & \textbf{0.8795}       & 1.6982        & 1.9346\\
MISO        &   $\times 10^{-2}$    & \textbf{4.1489}       & 5.8392        & 6.1908\\
PhysioNet   &   $\times 10^{-3}$    & \textbf{2.3266}       & 2.9804        & 3.4947\\
\bottomrule
\end{tabular}
\end{table}

\subsection{Settings}
We use Relative Mean Squared Error (RMSE) to evaluate the forecasting accuracy. We divide all datasets into training, validation, and test sets by a 70\%:10\%:20\% ratio. We present the results of the complete version of our model, i.e., a hybrid PDE model with the meta-learning controller, in the experimental tables. For both baselines and our model, we do grid search for the following hyperparameters, i.e., the learning rate \{1e-2, 3e-3, 1e-3\}, the batch size \{16, 32, 64, 128, 256\}, the hidden dimension \{16, 32, 64\}, the penalty weight on regularizers \{1e-5, 3e-5, 1e-4\}. Unique hyperparameters that are only used in specific baselines are set to the default numbers in their respective papers. Unique hyperparameters in our model are also obtained by grid search, i.e., $N_c$, $N_l$. The model with the best validation performance is applied for obtaining the results on the test set. To be noted, the three hyperparameters when controlled by the meta-learning model, i.e., inner coefficients $\epsilon$, sampling rate $r$, and time span $s$, are not obtained by grid search. Only in the ablation study, the three hyperparameters are obtained via grid search for the "hybrid" model. We perform both multi-step and single-step prediction experiments, where we set {50, 3600, 5, 20} as the number of steps for Synthetic, Orderbook, MISO, and PhysioNet datasets in multi-step experiments, respectively. 

\subsection{Performance Comparison}
In this subsection, we conduct performance comparison on different baselines models for both multi-step time series forecasting and single-step time series forecasting. 

First, Table \ref{tab:main1} shows the performance comparison with a multi-step forecasting setting.
The overall results show that our proposed learning differential operators have comparable performance to most baseline models. Neural ODE, and ODE-RNN with advanced architectures or optimization methods generally perform better than LSTM, ARIMA, and Prophet. ConvTrans and DeepAR further outperform Neural ODE and ODE-RNN due to the larger model capacity of the transformer-based structure. The advantage of our model is relatively more obvious in single-step forecasting setting than in multi-step setting, probably due to the fact that the Euler time-stepping always accumulates integration and estimation error at every forecasting step. The estimation error is inevitable since we use the derivatives to a previous data point to approximate the derivatives (variation slope) at the current data point, which is a delayed estimation. It is a compromise since the derivative at the current time is often not available. Through the integration of every time, this error is accumulated, making it hard for PDE models to predict too many future steps.

We find that the performance difference on different datasets varies a lot. For example, on the Synthetic dataset where the time series are generally smooth with only several waves as shown in Appendix, the performance differences between multi-step and single-step and the difference between different models are both relatively tiny. Our model slightly outperforms other models on Synthetic data. On the Orderbook dataset where the time series is also relatively smooth due to the moving average of the original time series, our model shows overwhelmingly higher accuracy over other models, especially with the single-step forecasting setting. The MISO dataset and PhysioNet dataset are more challenging to model, due to their irregular and/or unsmooth data samples. On MISO and PhysioNet, our model presents a comparable performance to the state-of-the-art ConvTrans model with a multi-step setting, which demonstrates its good performance even when the data is of low quality. With a single-step forecasting setting, our model can slightly outperform other models, showing its advantage in precisely capturing the sequential dynamics and predicting the next-time variation. This aligns with our intuition that PDE is designed to describe the local variation at each time step. 

\subsection{Ablation Study}
We perform an ablation study to analyze the importance of the various components in our model. We analyze the ablation of meta-learning and hybrid PDE model in Table \ref{tab:main3}. The results unequivocally show that the proposed meta-learning controller for optimizing the hyperparameters in real-time is crucial for improving the performance of PDE models. Compared to other models as shown in Table \ref{tab:main1}, using a single PDE to model the entire time series can only obtain comparable or sometimes worse performance than LSTM or Prophet. This is understandable as the single PDE model can hardly reflect the ever-changing dynamics. The hybrid PDE model can effectively improve the performance of the single PDE model, as the model can flexibly and selectively leverage the most important information in the time series for prediction. For example, if the dynamics are always changing rapidly, the hybrid model can freely set the time span to be small in order to focus more on the recently emerging pattern instead of the old sequential pattern. The meta-learning PDE model further improves the performance of the hybrid PDE model by allowing the real-time optimization of hyperparameters at every time step. For example, a meta PDE model can flexibly set the time span to be small when the dynamic is changing and set it to be large to focus on the long-term trend when the dynamics are quite stable. The results show that the proposed model has a unique advantage in capturing the dynamic change to improve the model performance on complex time series.

\subsection{Equation Visualization}
In this subsection, we present the mathematical expression of our proposed PDE learning model, which contains understandable information regarding the correlations between the multiple variables. As shown in Table \ref{tab:main3}, we visualize the most weighted PDE in the hybrid PDE model for the last multi-step predictions of the four datasets. As the PDE model is always changing to reflect time-evolving dynamics, here we present exemplified PDEs trained on historical data to predict a certain future time point for each dataset. Due to the sparse regression, only the most important terms (up to four terms) in the PDE model are presented. Visualization on meta-learning is shown in Appendix.

For the synthetic dataset of wave polynomials, we find that the time series dynamics are determined by the second-order derivatives of the target variable to the two correlated variables. According to this, we can know that the acceleration of the target value changing with time is directly proportional to the acceleration changing with other variables. Knowing the second-order differentiation to other variables, we can easily infer its time series dynamics. 

We further present the results of the Orderbook dataset. From the two terms related to $\frac{\partial x}{\partial t}$ on the right-hand side, we can know that the futures ($y$) rate of change and the spot ($x$) rate of change are highly correlated. We can rely heavily on the variation of spot price to predict futures price. In addition, the rate of change for futures is also influenced by the time evolution itself and the differential relation between the two variables. 

Furthermore, we find that many factors play important roles in determining the rate of change for electricity loads, such as the acceleration of the load changing with price, the time derivatives of electricity price, and the differential relation between load and price. This aligns well with the experimental results that the RMSE on MISO has relatively higher RMSE than on other datasets, as our model tries to leverage every possible correlation to help reveal the dynamics of electricity load with limited observable variables.

The time series forecasting of respiration rate on the PhysioNet dataset as an example also provides many interpretable insights. We find that the time derivative of respiration rate (y) is highly correlated to the time derivatives of temperature ($x_1$), partial pressure of arterial $CO_2$ ($x_2$), and the differential correlation between respiration rate and arterial pH ($x_3$). This aligns with our clinical experiences that fever or respiratory alkalosis can be statistically correlated to rapid and/or deep breathing. 

The mathematically interpretable model provides us with a dynamical system to explain the data. While we can also tell from the attention scores from deep learning models that futures price and spot price may be correlated, there lacks an explicit equation to describe the dynamics for these scenarios. Our proposed method, however, can extract the explicit knowledge from data and directly use it for state-of-the-art forecasting performance. 

\begin{table}[t]
\centering
  \caption{The mathematical form of the major trend captured by PDE for the last multi-step prediction of each sequence.}
  \label{tab:main4}
\begin{tabular}{cc}
\toprule
Dataset & Partial Differential Equations\\
\midrule
Synthetic   &   $\frac{\partial^2 y}{\partial t^2} = 1.02 \frac{\partial^2 y}{\partial x_1^2} + 0.99 \frac{\partial^2 y}{\partial x_2^2}$\\
\midrule
\multirow{2}{*}{Orderbook}   &   $\frac{\partial y}{\partial t} = -0.067 \frac{\partial x}{\partial t} \cdot x + 0.044 \frac{\partial x}{\partial t}$\\ 
 & $+ -0.007 t + 0.003 \frac{\partial y}{\partial x} \cdot x$\\
\midrule
\multirow{2}{*}{MISO}        &   $\frac{\partial y}{\partial t} = -1.157 \frac{\partial^2 y}{\partial x^2} + 1.144 \frac{\partial^2 y}{\partial x^2} \cdot t$\\
 & $+ 0.062 \frac{\partial x}{\partial t} \cdot t - 0.032 \frac{\partial y}{\partial x} \cdot x$\\
 \midrule
\multirow{2}{*}{PhysioNet}   &   $\frac{\partial y}{\partial t} = -0.274 \frac{\partial x_1}{\partial t} + 0.091 \frac{\partial x_2}{\partial t} \cdot t$\\
 & $+ 0.088 \frac{\partial y}{\partial x_2} - 0.062 \frac{\partial y}{\partial x_3} \cdot x$\\
\bottomrule
\end{tabular}
\end{table}

\section{Conclusion}
In this paper, we introduce the learning of differential operators for interpretable time series forecasting. We propose a learnable novel $P$-block to capture the differential relations from data and use a recurrent differential model that mimics the Euler time stepping of the $P$-block for forecasting. This mechanism can successfully apply the modeling of high-dimensional dynamical systems for multivariate time series forecasting. In addition, we propose a hybrid PDE model that uses different proportions of data to balance the modeling of long-short dependencies. Accordingly, we further propose a meta-learning controller to form a dynamic PDE model, which can capture the change of sequential dynamics and adjust the hyperparameters of the hybrid model adaptively. Extensive experiments on synthetic and real-world datasets demonstrate that our model has comparable performance to state-of-the-art deep learning models and can provide interpretable mathematical expressions for describing the dynamics. The interpretability of a concise PDE roots on the assumption that most sequential patterns are governed by a smooth major trend. In the future, how to collaboratively use other suitable models for capturing unsmooth minor variations is important for further enhancing the balance between interpretability and accuracy.

\bibliographystyle{ACM-Reference-Format}
\balance
\bibliography{ref}

\appendix
\section{APPENDIX}
\subsection{Algorithm Details}
The implementation procedures of the proposed algorithm is shown in Algorithm \ref{alg_training}.
\begin{algorithm}
    \caption{The training process of dynamic PDE model.}
    \label{alg_training}
    \begin{algorithmic}[1]
        \REQUIRE Multivariate time series $x$, target time series $y$, time stamps of time series $t$, number of points $m$, hybrid PDE $H$, meta-learning controller $G$, number of hyperparameter set $n$. 
        \ENSURE Model parameters $\theta$ for $G$ and $\omega$ for $H$.
        \STATE Initialize model parameters $\theta$ and $\omega$ randomly.
        \STATE Initialize i = 1, p = 0, m' = 70\% m.
        \WHILE{i < m'.}
            \WHILE{p < n.}
                \STATE Generate candidate hyperparameters $s, r, \epsilon$, train $H$ based on $x, y, t, s, r, \epsilon$, obtain the optimal estimation error of $H$.
                \STATE $p \gets p + 1$
            \ENDWHILE
            \STATE Initialize $p = 0$. Use the estimation errors of $H$ under different hyperparameters and $x, y, t, s, r, \epsilon$ to train $G$.
            \STATE $i \gets i+1$.
        \ENDWHILE
        \RETURN $H$ and $G$.
    \end{algorithmic}
\end{algorithm}

\subsection{Synthetic Data Visualization}
\begin{figure}[h]
\centering
\includegraphics[width=0.45\textwidth]{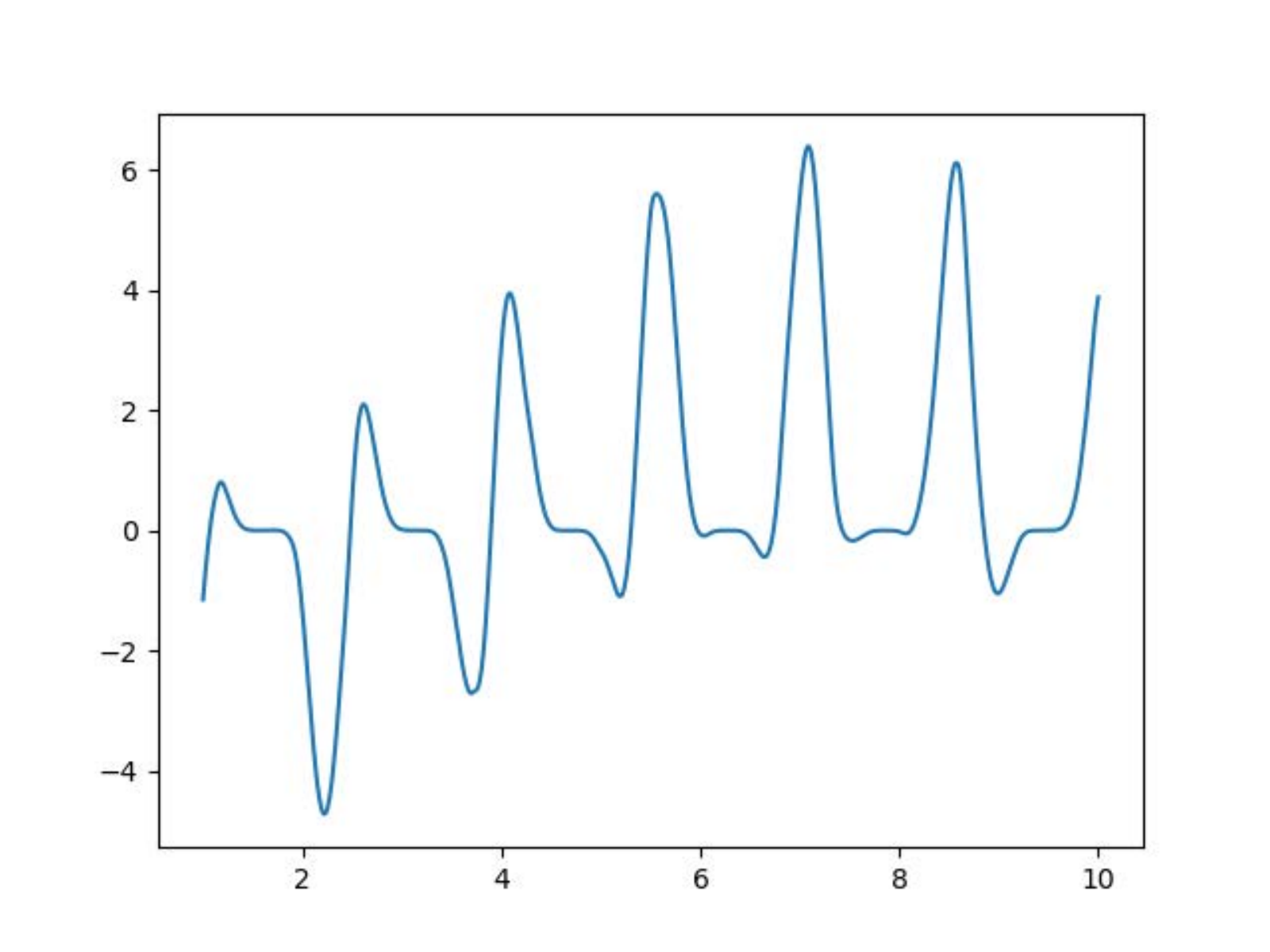}
\caption{An example time series synthesized by Eq. 12. The abscissa represents time, and the ordinate represents value.}
\label{syn}
\end{figure}

We visualize the synthetic data in Figure \ref{syn}. It shows multiple waves, as the data is generated by a special solution (with certain boundary conditions and initial conditions) to a wave equation in physics.

\subsection{Meta-Learning Visualization}
In this subsection, we further show examples that the proposed meta-learning mechanism captures changing dynamics for time series forecasting. We visualize some exemplified dynamic stages discovered by the proposed meta-learning PDE models on the Orderbook dataset in Table \ref{tab:meta}. Each stage indicates a different time with different sequential patterns. The most weighted PDE in the hybrid PDE model for the forecasting of different times is presented for each stage. The ranking of terms is determined by the contributions, i.e. the absolute value of a coefficient times the term. Therefore, the first term has a larger influence on the equation than the second term. Note that the third stage uses the same result (from the last time prediction) presented in Table \ref{tab:main4}. 

We find that the time derivative of the futures price of the first stage is not only correlated to the time derivative of spot prices as in the third stage but is also determined by the futures price itself and a bias term. The first stage is exactly at the start of a trading quarter, which may explain the appearance of linear terms like $y$ and $1$ as an extra adjustment to the previous quarter of the market. The second stage is in the middle of a trading quarter, where we find that the time derivative of spot prices has an overwhelming influence on the variation of futures prices than other terms. This may suggest a smooth stage where futures and spot prices are highly correlated. The third stage, on the other hand, represents a less special case where the model may exploit many potentially useful terms to co-determine the complex time-series dynamics. Because our meta-learning model can capture the dynamics change from the data, it is flexible to obtain the major trend of the ever-changing dynamics with mathematical interpretations.

\begin{table}[t]
\centering
  \caption{The different time series dynamics at different times of Orderbook dataset captured by the meta-learning model.}
  \label{tab:meta}
\begin{tabular}{cc}
\toprule
Stages & Partial Differential Equations\\
\midrule
\multirow{2}{*}{First}   &   $\frac{\partial y}{\partial t} = 4.224 \times 10^{-7} y - 0.186$ \\ 
 & $+ 1.632 \frac{\partial x}{\partial t} - 0.023 \frac{\partial x}{\partial t} \cdot t$\\
\midrule
\multirow{2}{*}{Second}   &   $\frac{\partial y}{\partial t} = 2.134 \frac{\partial x}{\partial t} - 1.562 \times 10^{-6} \frac{\partial x}{\partial t} \cdot x$\\
 & $- 7.720 \times 10^{-7} \frac{\partial x}{\partial t} \cdot t$\\
 \midrule
\multirow{2}{*}{Third}  &   $\frac{\partial y}{\partial t} = -0.067 \frac{\partial x}{\partial t} \cdot x + 0.044 \frac{\partial x}{\partial t}$\\ 
 & $+ -0.007 t + 0.003 \frac{\partial y}{\partial x} \cdot x$\\
\bottomrule
\end{tabular}
\end{table}

\end{document}